\documentclass[letterpaper, 10 pt, conference]{ieee_format/ieeeconf}  

\IEEEoverridecommandlockouts                              

\usepackage[noadjust]{cite}
\usepackage{threeparttable}
\usepackage{url}
\usepackage{xcolor}
\usepackage{kotex}

\usepackage{graphicx}
\usepackage{amssymb} 
\usepackage{amsmath}

\usepackage{siunitx}
\usepackage{float}
\usepackage{dsfont}
\usepackage{multicol}
\usepackage{bm}
\usepackage{booktabs}
\usepackage{tabularx}
\let\labelindent\relax
\usepackage{enumitem}

\usepackage{array}       
\usepackage{tabularx}    
\usepackage{booktabs}    
\usepackage{multirow}   
\usepackage{amsmath}     
\usepackage{bm}          
\usepackage{bbm}         
\usepackage{threeparttable}
\usepackage{caption}     
\usepackage{makecell}

\usepackage{multirow}

\usepackage{caption}
\title{\LARGE \bf
Phase-Aware Policy Learning for Skateboard Riding

of Quadruped Robots via Feature-wise Linear Modulation
}

\author{
Minsung Yoon\textsuperscript{*}, 
Jeil Jeong\textsuperscript{*},
Sung-Eui Yoon
}

\begin{document}
    \maketitle
    \thispagestyle{empty}
    \pagestyle{empty}

    \begin{abstract}
Skateboards offer a compact and efficient means of transportation as a type of personal mobility device. 
However, controlling them with legged robots poses several challenges for policy learning due to perception-driven interactions and multi-modal control objectives across distinct skateboarding phases.
To address these challenges, we introduce Phase-Aware Policy Learning (\textit{PAPL}), a reinforcement-learning framework tailored for skateboarding with quadruped robots. 
\textit{PAPL} leverages the cyclic nature of skateboarding by integrating phase-conditioned Feature-wise Linear Modulation layers into actor and critic networks, enabling a unified policy that captures phase-dependent behaviors while sharing robot-specific knowledge across phases. 
Our evaluations in simulation validate command-tracking accuracy and conduct ablation studies quantifying each component’s contribution. We also compare locomotion efficiency against leg and wheel–leg baselines and show the real-world transferability.

\end{abstract}
    \section{Introduction}
Legged robots have shown robust locomotion across challenging terrains, including icy, rough, and deformable surfaces~\cite{arm2023scientific, lindqvist2022multimodality, bellicoso2018advances}. 
However, their leg-based actuation inherently limits speed and energy efficiency, especially for long-range missions under constrained battery capacity~\cite{hwangbo2025raibo2}. 
To mitigate this, recent studies have investigated augmenting robots with riding capabilities~\cite{thibault2024learning, takasugi20173d, takasugi2018extended, kimura2018riding, anglingdarma2021motion, xin2017torque, rajendran2022towards, gong2019feedback, xu2024optimization, liu2025discrete}. 
This modality allows the robot to leverage personal mobility devices, such as skateboards or Segways, enabling more efficient long-distance travel while conserving onboard energy, analogous to human behavior.

Among these devices, the skateboard is a lightweight, non-motorized platform that offers distinct advantages for legged robots without manipulators. 
Propulsion arises from forces applied to the deck---whether from the rider’s kicking impact or gravity---which are converted into wheel torques through wheel–ground interaction, while the low rolling resistance of the bearings minimizes energy loss. 
Furthermore, the skateboard’s truck assembly allows steering via weight shifting on the deck, eliminating the need for grasp-based manipulation. 
Sec.~\ref{sec:skateboard_dynamics} provides a detailed description of these mechanisms.

Despite these advantages, the design of skateboard-riding controllers for legged robots remains challenging. 
The skateboarding motion is multi-modal, comprising cyclic pushing and carving phases, and transitions where the feet designated for propulsion alternate between the ground and the board.
Each phase entails distinct contact patterns, dynamics, and control objectives, with seamless transitions further complicating the design of a unified controller.
In addition, the robot must sustain balance against non-inertial forces induced by the board’s acceleration and avoid desynchronization caused by prolonged ground contact.
Moreover, persistent stability requires exteroceptive–control coupling for synchronization and recovery from unstable states through posture adaptation.

\begin{figure}[t]
    \vspace{0.2cm}
    \centering 
    \includegraphics[width=\columnwidth]{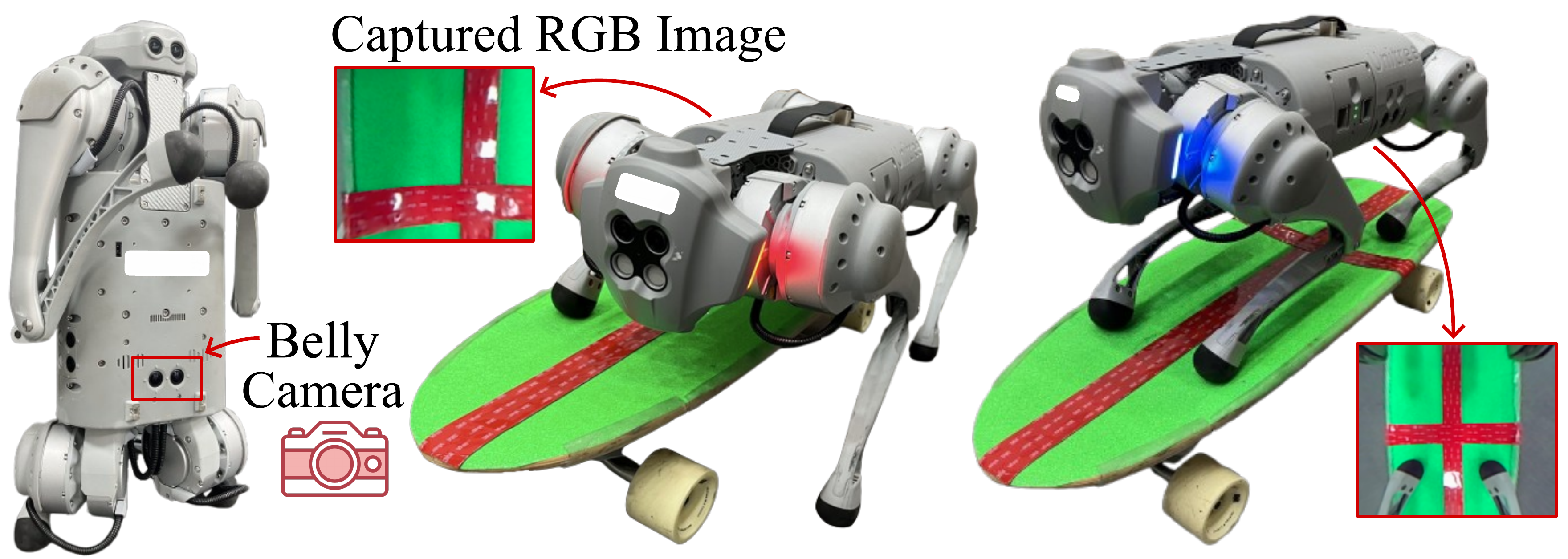}
    \vspace{-0.5cm}
    \caption{
        Belly-mounted RGB camera setup on the Unitree Go1 robot~\cite{Unitree_Go1}. The camera observes the skateboard deck surface and supplies visual feedback for localization and control, enabling resilient skateboarding maneuvers.
    }
    \label{fig:main_fig}
    \vspace{-0.7cm}
\end{figure}

Recent studies have applied model-based optimization~\cite{xu2024optimization} and Reinforcement Learning (RL)~\cite{liu2025discrete} for skateboarding of quadruped robots.
The model-based method relies on a pre-computed motion library and a linearized model for tracking, but limited model fidelity and library coverage could reduce robustness to unforeseen events such as slippage.
In contrast, the prior RL-based research develops policies directly from interaction, but it simplifies the training setup by assuming a foot–board attachment in simulation, reducing the underactuated dynamics of two floating bases to a coupled system.
While this design choice facilitates training, it could reduce generalization at deployment and restrict maneuvers mainly to gliding rather than steering through deck tilting.
In this work, we aim to relax such assumptions and integrate exteroceptive perception to enable sustained skateboarding with visual feedback, as illustrated in Fig.~\ref{fig:main_fig}.
To the best of our knowledge, this represents one of the first efforts to utilize onboard exteroceptive sensing for quadruped skateboarding.

We propose Phase-Aware Policy Learning (\textit{PAPL}), an RL framework that leverages the cyclic and multi-modal nature of skateboard riding. 
\textit{PAPL} incorporates a phase-conditioned modulation mechanism within actor–critic networks, effectively encoding phase-specific behaviors in a unified policy.
To induce proficient riding skills under partial observability, we adopt asymmetric privileged learning followed by a distillation process as a two-stage learning scheme.
This improves sample efficiency by exploiting informative privileged states in simulation and subsequently replacing them with estimates from observation history.
In particular, exploiting and inferring the relative board states with respect to the robot enables real-time posture adaptation for resilient skateboarding.

In our experiments, we evaluate the command-tracking accuracy of our skateboard-riding policy and conduct ablation studies to reveal the contributions of individual components within \textit{PAPL}. 
We also assess locomotion efficiency relative to legged and wheel–legged baselines and examine sim-to-real transferability under diverse environmental conditions.

    \section{Variable Notation}
\begin{figure}[t!]
    \centering 
    \includegraphics[width=\columnwidth]{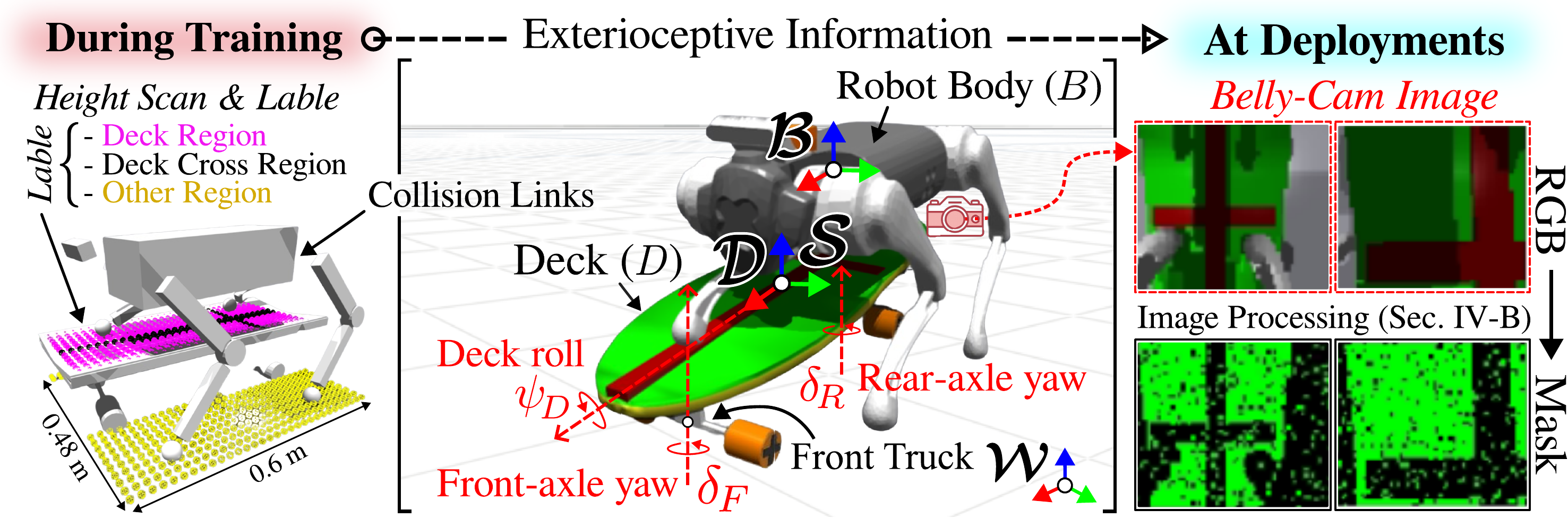}
    \caption{ 
    The center shows the reference frames: robot body $\mathcal{B}$, skateboard $\mathcal{S}$, deck $\mathcal{D}$, and world $\mathcal{W}$, along with the physical robot body $B$ and deck $D$ objects. The skateboard frame $\mathcal{S}$ is fixed to the board, with the deck $\mathcal{D}$ rotating about its roll axis relative to $\mathcal{S}$; joint variables include deck roll ($\psi_D$) and the yaw angles of the front and rear wheel axles ($\delta_F, \delta_R$). 
    The left and right show exteroceptive inputs for training and deployment stages.
    }
    \vspace{-5.5mm}
    \label{fig:vars}
\end{figure}

In Cartesian space, translational variables such as position $\mathbf{p}$, linear velocity $\mathbf{v}$, acceleration $\dot{\mathbf{v}}$, and force $\mathbf{f}$, as well as rotational variables including XYZ Euler angles $\boldsymbol{\theta}$, angular velocity $\boldsymbol{\omega}$, angular acceleration $\dot{\boldsymbol{\omega}}$, and torque $\boldsymbol{\tau}$, are defined in $\mathbb{R}^3$. 
For clarity, we employ superscripts to denote reference frames and subscripts to specify the object, coordinate, and time index. 
For instance, $p^{\mathcal{S}}_{B,x,t{-}1}$ denotes the $x$-component of the body’s position $p_B$ represented in the skateboard frame $\mathcal{S}$ at time $t{-}1$. 
For brevity, we omit the current time index $t$. 
For quadruped robots with 12 actuated joints, joint position $\boldsymbol{q}$, velocity $\dot{\boldsymbol{q}}$, acceleration $\ddot{\boldsymbol{q}}$, and torque $\boldsymbol{\tau}_q$ lie in $\mathbb{R}^{12}$. 
Each foot $F_i$, where $i \in \{0,1,2,3\}$, is associated with a contact force $\mathbf{f}_{F_i}$ and binary contact state $c_i \in \{0,1\}$. 
We define $\mathcal{F}_{\text{all}}$, $\mathcal{F}_{\text{left}}$, and $\mathcal{F}_{\text{right}}$ as the index sets of all feet, left-side feet, and right-side feet, respectively, and $\mathcal{J}_{\text{all}}$, $\mathcal{J}_{\text{left}}$, and $\mathcal{J}_{\text{right}}$ as the corresponding index sets of actuated joints, grouped by leg.
For skateboards, representative joints include the deck roll angle $\psi_D$ and yaw angles of the front and rear wheel axes, $\delta_F$ and $\delta_R$, all in $\mathbb{R}$, as illustrated in Fig.~\ref{fig:vars}.

\section{Skateboard Dynamics Modeling} \label{sec:skateboard_dynamics}
Accurate modeling of skateboard dynamics is essential to develop riding policies and narrow the sim-to-real gap.
Thus, our model captures two mechanisms: steering, resulting from the truck's trigonometric geometry; and propulsion, which is inherently passive but modeled by applying wheel torques to compensate for contact modeling limitations in simulation.

\subsection{Steering Dynamics}
The skateboard is a passive, non-holonomic system with a rigid deck mounted on front and rear trucks, each connecting a pair of wheels via an axle.
Steering arises from a roll-to-yaw coupling: rider-induced deck roll is converted into yaw rotation of the wheel axles.
This coupling defines a nonlinear relationship between the deck roll angle $\psi_D$ and the yaw angles of the front and rear axles, $\delta_F$ and $\delta_R$, expressed as:
\begin{equation}
\delta_i = \arctan\left( \gamma^1_i \sin\left( \gamma^2_i \psi_D \right) \right), \; i \in \{F, R\}
\label{eq:steering}
\end{equation}
where $\gamma^1_i$ and $\gamma^2_i$ are skateboard parameters determined by the mechanical configuration~\cite{dyn2}.
We emulate this passive steering mechanism using Proportional-Derivative (PD) controllers at each axle to track the desired angles $\delta_F$ and $\delta_R$.

Thus, the rider achieves effective steering by actively modulating the deck roll $\psi_D$ through contact force distribution on the deck. 
Its roll dynamics, governed by bushing compliance and rider-induced torques, is modeled as follows:
\[
    \boldsymbol{\tau}^{\mathcal{D}}_{D} = \sum_{i \in \mathcal{F}_{\text{is-on-deck}}} \mathbf{p}^{\mathcal{D}}_{F_i} \times \mathbf{f}^{\mathcal{D}}_{F_i} ,
    \label{eq:roll_contact_torque}
\]
\[
    \dot{\omega}^{\mathcal{D}}_{D, \text{x}} = (-k_{\text{bushing}}^\text{P} \psi_D - k_{\text{bushing}}^\text{D} \dot{\psi}^D + \tau^{\mathcal{D}}_{D, \text{x}}) / I_{D, \text{xx}},
    \label{eq:roll_angular_accel}
\]
where $k_{\text{bushing}}^\text{P}$ and $k_{\text{bushing}}^\text{D}$ represent the bushing stiffness and damping coefficients of the restoring torque, and $I_{D,\text{xx}}$ is the deck's moment of inertia about its roll axis.

\subsection{Propulsion Dynamics}
Physics engines often exhibit limited accuracy in modeling wheel–ground contact and rolling resistance~\cite{liu2025discrete}.
This inaccuracy prevents external forces applied to the deck from being fully transmitted to the wheels as effective driving torque, resulting in unrealistic propulsion behavior.
To mitigate this issue, we explicitly compute and apply effective torques to the wheels.
The net torque applied to left and right wheels $j \in \{L, R\}$ of front and rear axles $i \in \{F, R\}$ is defined as
\[
\tau_{i, j}^{\text{net}} = \tau_{i}^{\text{ext}} - \tau_{i, j}^{\text{fric}}.
\]

The external torque $\tau_i^{\text{ext}}$ is computed by projecting the net force acting on the deck along the rolling direction of each axle. 
The net force in the skateboard frame $\mathcal{S}$ is modeled as
\[
\mathbf{f}_{D}^{\mathcal{S}} = -\sum_{i \in \mathcal{F}_D} \mathbf{f}_{F_i}^{\mathcal{S}} + m_S \mathbf{g}^{\mathcal{S}},
\]
where $m_S$ is the skateboard mass and $\mathbf{g}$ is gravitational acceleration. 
Due to non-holonomic constraints, only the force aligned with the axle directions contributes to propulsion. 
The axle direction is represented by the unit vector $\hat{\boldsymbol{d}}_i^{\mathcal{S}} = (\cos \delta_i,\ \sin \delta_i,\ 0)^\top$, with the corresponding projected force $f^{\text{drive}}_i = \boldsymbol{f}_{D}^{\mathcal{S}} \cdot \hat{\boldsymbol{d}}_i^{\mathcal{S}}$ at each axle $i$. 
Assuming symmetric torque distribution across wheels, we compute the resulting external torque as $\tau_i^{\text{ext}} = \frac{r}{2} f^{\text{drive}}_i$, where $r$ is the wheel radius.

The friction torque $\tau_{i, j}^{\text{fric}}$ accounts for static friction as well as dynamic effects---including viscous damping and rolling resistance---as a function of the wheel's angular velocity $\omega_{i,j}$:
\[
\tau_{i, j}^{\text{fric}} = \!
\begin{cases}
\min\left( \left| \tau_{i}^{\text{ext}} \right|,\ \tau_{\text{static}} \right) \cdot \operatorname{sign}\left( \tau_{i}^{\text{ext}} \right), \!\!& \!\!\! |\omega_{i, j}| < \epsilon \\
c_\omega \omega_{i, j} + r \mu_r m_S ||\mathbf{g}||_2 \cdot \operatorname{sign}(\omega_{i, j}), \!\!& \!\!\! |\omega_{i, j}| \geq \epsilon
\end{cases},
\]
where $\tau_{\text{static}}$ is the static friction limit, $\mu_r$ is the rolling resistance coefficient, and $c_\omega$ is the viscous damping constant.

\begin{figure}[t!]
    \centering 
    \includegraphics[width=\columnwidth]{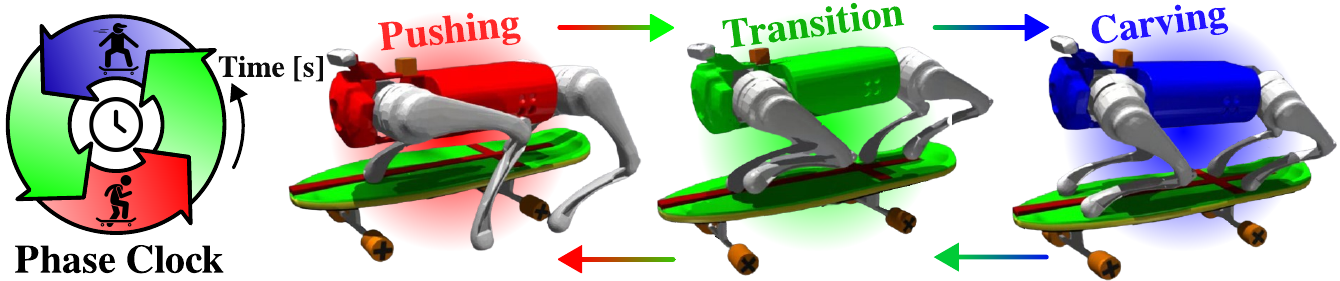}
    \caption{ 
    Illustration of the phase clock concept that manages the cyclic nature of skateboarding, along with representative motion snapshots of each phase.
    }
    \vspace{-0.55cm}
    \label{fig:phase_clock}
\end{figure}

\begin{figure*}[t]
    \centering 
    \includegraphics[width=2.0\columnwidth]{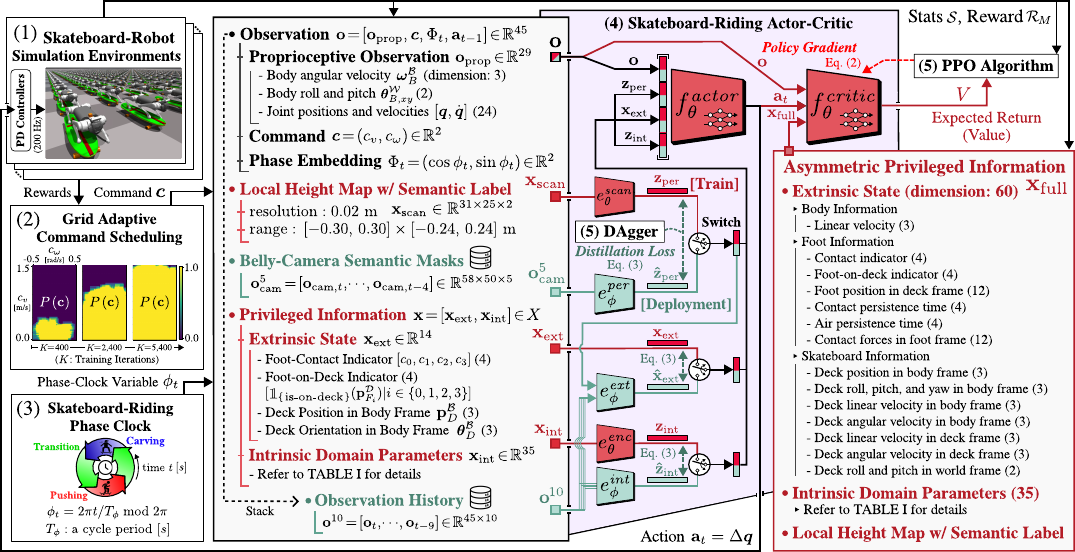}
    \caption{
\textbf{Phase-Aware Policy Learning (\textit{PAPL}) Framework for Skateboard Riding.}
(1) Simulation environments modeling skateboard–robot interaction.
(2) Command scheduling that procedurally increases riding difficulty for broad command-space coverage~\cite{margolis2024rapid}.
(3) Phase-clock representation that alternates over time between pushing, transition, and carving modes.
(4) An asymmetric actor–critic architecture: the critic leverages full privileged information for effective policy guidance with clear situational awareness, while the actor relies solely on features that can be inferred or directly observed.
(5) Proximal Policy Optimization (PPO)~\cite{schulman2017proximal} trains policy networks parameterized by $\theta$ (red-color networks) to maximize Eq.~(\ref{eq:objective_function}). The converged policy is then distilled via Dataset Aggregation (DAgger)~\cite{ross2011reduction} for the estimators parameterized by $\phi$ (mint) using Eq.~(\ref{eq:dagger}), replacing inaccessible information during deployment. 
    }
    \vspace{-0.55cm}
    \label{fig:overall}
\end{figure*}

    \section{Skateboard-Riding Policy Learning} \label{sec:method}
We present the Phase-Aware Policy Learning framework (\textit{PAPL}), which enables quadruped robots to ride skateboards.
To exploit the riding task's cyclic nature shown in Fig.~\ref{fig:phase_clock}, we integrate a phase clock into the learning process and policy architecture.
The following description provides the problem formulation, policy composition, and implementation details.

\subsection{Formulation of Skateboarding Policy Learning}
\label{Method-A}
We formulate a skateboarding task as a phase-conditioned reinforcement learning problem, where the quadruped robot learns to perform riding skills---pushing, mounting, carving, and foot planting---on a dynamic skateboard, as shown in Fig.~\ref{fig:phase_clock}.
To represent the riding task’s cyclic and multi-modal structure, we introduce a phase clock variable \(\phi_t \in [0, 2\pi)\):
\[
\phi_t = 2\pi t/T_\phi \bmod 2\pi,
\]
where \(T_\phi\) denotes the period of one skateboarding cycle. This cyclic phase variable determines a discrete motion mode \(M\):
\[
M\hspace{-0.08em}(\phi_t) =
\begin{cases}
\textsc{Carving} & \text{if } \phi_t \in [0.2\pi,\ 0.8\pi], \\
\textsc{Pushing} & \text{if } \phi_t \in [1.2\pi,\ 1.8\pi], \\
\textsc{Transition} & \text{otherwise}.
\end{cases}
\]

We employ the phase variable \(\phi_t\) to modulate distinct control intents of the policy.
Accordingly, we model the task as a phase-conditioned Partially Observable Markov decision process (POMDP), defined by the tuple \((\mathcal{S}, \mathcal{O}, \mathcal{A}, \mathcal{T}, \mathcal{R}_{\hspace{-0.1em}M}, \rho_0, \gamma)\), where \(\mathcal{S}\) is the state space, \(\mathcal{O} \subset \mathcal{S}\) the observation space, \(\mathcal{A}\) the action space, \(\mathcal{T}\) the transition dynamics, \(\mathcal{R}_{\hspace{-0.1em}M}\) the mode-dependent reward function, \(\rho_0\) the initial state distribution, and \(\gamma\) the discount factor. 
We set the initial state \(\boldsymbol{s}_0 \sim \rho_0\) by placing the robot at the center of the skateboard in a nominal crouching posture \(\boldsymbol{q}_0\), with small deviations in \(\mathbf{p}^{\mathcal{S}}_{B}\) and \(\boldsymbol{q}\).
We then optimize the skateboard-riding policy $\pi_\theta$ by maximizing the expected return over commands  $\boldsymbol{c}$ and phase periods \(T_\phi\):
\begin{equation}
\hspace{-0.45em}J(\theta) \! = \!
\mathbb{E}_{\!\substack{\boldsymbol{c} \sim P(\boldsymbol{c}), \\ T_\phi \sim P(T_\phi\hspace{-0.05em})}} \!\!
\left[
\mathbb{E}_{\!\! \substack{ \boldsymbol{s}_0 \sim \rho_0, \\ (\boldsymbol{s}, \boldsymbol{a}) \sim \rho_{\pi_\theta}}} \hspace{-0.45em}
\left[ 
\sum_{t=0}^{\infty} \gamma^{t} \hspace{0.05em}\mathcal{R}_{\hspace{-0.1em}M\hspace{-0.08em}(\phi_t\hspace{-0.05em})}\hspace{-0.08em}(\boldsymbol{s}_t, \!\boldsymbol{a}_t \!\!\mid \!\boldsymbol{c})
\right]
\right]\!,
\label{eq:objective_function}
\!\!\!
\end{equation}
where $\rho_{\pi_\theta}$ is the state-action visitation distribution under the policy $\pi$ parameterized by $\theta$, and $\boldsymbol{c} = (c_v, c_\omega) \in \mathbb{R}^2$ denotes a set of forward velocity and yaw rate commands. 
To facilitate skill acquisition, we employ a grid-adaptive scheduling strategy over the command distribution \(P(\boldsymbol{c})\)~\cite{margolis2024rapid}. As shown in Fig.~\ref{fig:overall}-(2), task difficulty is progressively increased by expanding command space in line with policy proficiency. For the phase-period distribution \(P(T_\phi)\), we uniformly sample \(T_\phi \sim \mathcal{U}(4.0, 12.0)~\SI{}{\second}\) at the beginning of each episode.

Partial observability in POMDPs introduces challenges in learning complex motor skills through direct policy optimization~\cite{gangwani2020learning, meng2021memory, fu2023deep}. 
To address this, we adopt a privileged learning framework that reformulates the problem as an MDP using privileged information \( X \subset S \setminus O \)~\cite{Yu2017si, lee2020learning, miki2022learning, cheng2024extreme, kumar2021rma, kumar2022adapting, yoon2025enhancing}.
This privileged state information $X$ provides richer environmental context to the policy, facilitating optimization with respect to Eq.~(\ref{eq:objective_function}).

Specifically, we adopt a two-stage learning procedure that integrates privileged learning with system identification~\cite{kumar2021rma, kumar2022adapting, cheng2024extreme}. 
In the first stage, we develop a skateboard-riding policy using Proximal Policy Optimization (PPO)~\cite{schulman2017proximal} along with an asymmetric actor-critic architecture: the critic utilizes the full privileged state available in simulation, while the actor exploits a subset that is inferable from sensor observations.
In the second stage, we replace the actor’s simulation-only components with estimates to enable policy deployment under partial observability.  
To this end, we train three complementary estimators using the Dataset Aggregation (DAgger) algorithm~\cite{ross2011reduction}. 
As shown in Fig.~\ref{fig:overall}-(4), the estimators infer inaccessible privileged features from observation history and provide surrogate inputs to the actor network at deployment.

\begin{figure}[t!]
    \centering 
    \includegraphics[width=\columnwidth]{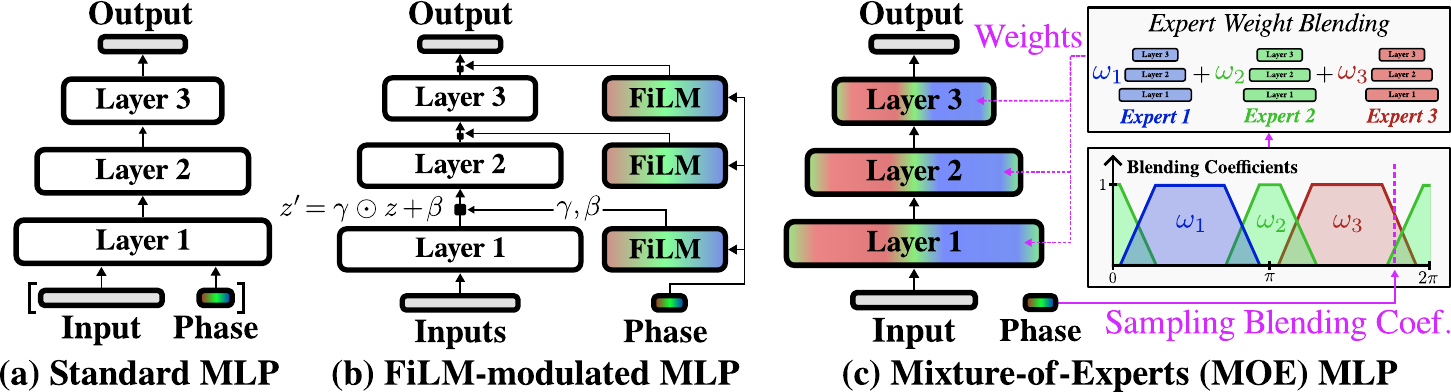}
    \caption{
        \textbf{Multilayer perceptron (MLP) network variants.} (a) Standard MLP, (b) FiLM-modulated MLP with phase-conditioned feature-wise modulation, and (c) Mixture-of-Experts MLP with phase-based expert weight blending.
    }
    \label{fig:NNs}
    \vspace{-0.6cm}
\end{figure}

\subsection{Phase-Aware Policy Composition} \label{sec:arch}
To encode phase-dependent multi-modal behaviors within a unified policy, we compose the actor $f_\theta^{\textit{actor}}$ and critic $f_\theta^{\textit{critic}}$ networks with Feature-wise Linear Modulation (FiLM)~\cite{perez2018film, 10160851, filmdiffusion}, which serves as an effective structural inductive bias for handling strong modality shifts in input–output distributions conditioned on specific variables.
As illustrated in Fig.~\ref{fig:NNs}-(b), we implement each layer in the FiLM-modulated multilayer perceptron (MLP) $\mathcal{F}$ to apply an affine transformation followed by modulation conditioned on the phase embedding $\Phi_t = (\cos\phi_t, \sin\phi_t)^\top \in \mathbb{R}^2$.
The computation at layer $\ell$ is:
\[
z = \mathbf{W}_\ell \mathbf{h}_{\ell-1} + \mathbf{b}_\ell, \quad
\mathbf{h}_\ell = \sigma\left( \gamma_\ell(\Phi_t) \odot z + \beta_\ell(\Phi_t) \right)
\]
where $\mathbf{h}_\ell$, $\mathbf{W}_\ell$, and $\mathbf{b}_\ell$ denote the layer activation, weight, and bias; $\gamma_\ell(\Phi_t)$ and $\beta_\ell(\Phi_t)$ are phase-conditioned modulation parameters; $\sigma(\cdot)$ is a nonlinear activation function; and $\odot$ is element-wise multiplication.
This modulation enables the networks to adapt to phase-specific variations while sharing robot-specific knowledge, facilitating smooth transitions and efficient motor skill reuse across the skateboard-riding cycle.

At each time step, the actor network $f_\theta^{\textit{actor}}$ outputs joint-displacement actions $\Delta \boldsymbol{q} \in \mathcal{A}$, representing deviations from the nominal posture $\boldsymbol{q}_0$.
These are converted into torques $\boldsymbol{\tau}_q$ by applying $\boldsymbol{q}_0 + \Delta \boldsymbol{q}$ as targets to the joint PD controllers.
The actor takes as input a structured tuple $(\mathbf{o}, \mathbf{z}_{\text{per}}, \mathbf{x}_{\text{ext}}, \mathbf{z}_{\text{int}})$, where 
the observation $\mathbf{o}$ includes proprioceptive information $\mathbf{o}_{\text{prop}}$, the command $\boldsymbol{c}$, the phase embedding $\Phi_t$, and the previous action $\boldsymbol{a}_{t{-}1}$; 
the perceptual feature $\mathbf{z}_{\text{per}}$ is produced by a scan encoder $e_\theta^{\textit{scan}}\!: \mathbf{x}_{\text{scan}} \! \rightarrow  \mathbf{z}_{\text{per}}$,  capturing skateboard-aware local height and semantic information, as depicted in Fig.~\ref{fig:vars}; 
the extrinsic privileged state $\mathbf{x}_{\text{ext}}$ consists of the deck's pose relative to the robot body as well as foot information; 
and the intrinsic latent feature $\mathbf{z}_{\text{int}}$ is generated by the encoder $e_\theta^{\textit{enc}}\!: \mathbf{x}_{\text{int}} \! \rightarrow  \mathbf{z}_{\text{int}}$.
Intrinsic parameters $\mathbf{x}_{\text{int}}$ (TABLE~\ref{table:domainparams}) affect the transition dynamics and contribute to the domain gap.
\begin{table}[b]
\scriptsize
\vspace{-0.35cm}
\centering
\renewcommand{\arraystretch}{1.1}
\setlength{\tabcolsep}{4pt}
\begin{tabular}{p{2.3cm}
    >{\centering\arraybackslash}p{0.47cm}  
    c 
    c 
    >{\arraybackslash}p{1.13cm}}
\hline
\textbf{Term} & \textbf{Dim.} & \textbf{Training Range} & \textbf{Testing Range} & \textbf{Unit} \\
\hline
\multicolumn{5}{c}{\textit{Quadruped Robots}} \\
\hline
Payload Mass               & (1)  & $[0.0,\,1.5]$       & $[0.0,\,3.0]$         & \si{\kilogram} \\
Shifted CoM                & (3)  & \hspace{-6.2pt}$[-0.05,\,0.05]$      & \hspace{-6pt}$[-0.1,\,0.1]$        & \si{\meter} \\
Friction Coef.              & (1)  & $[0.8,\,1.2]$       & $[0.7,\,2.0]$         & - \\
Restitution Coef.           & (1)  & $[0.0,\,0.1]$       & \hspace{4.0pt}$[0.0,\,0.15]$         & - \\
Leg-Joint PD Stiffness         & (12) & $[36.0,\,44.0]$         & $[34.0,\,46.0]$           & \si{\newton\meter}\,/\,\si{\radian} \\
Leg-Joint PD Damping           & (12) & $[0.8,\,1.2]$       & $[0.7,\,1.3]$         & \si{\newton\meter\second}\,/\,\si{\radian} \\
\hline
\multicolumn{5}{c}{\textit{Skateboards}} \\
\hline
Deck Mass                   & (1)  & $[3.5,\,4.5]$       & $[3.0,\,5.0]$         & \si{\kilogram} \\
Truck-Yaw PD Stiff.   & (1)  & $[45.0,\,50.0]$         & $[40.0,\,55.0]$           & \si{\newton\meter}\,/\,\si{\radian} \\
Truck-Yaw PD Damp.     & (1)  & $[2.5,\,3.0]$       & $[2.2,\,3.3]$         & \si{\newton\meter\second}\,/\,\si{\radian} \\
Bushing PD Stiffness & (1)  & $[30.0,\,35.0]$       & $[25.0,\,40.0]$         & \si{\newton\meter}\,/\,\si{\radian} \\
Bushing PD Damping  & (1)  & $[1.8,\,2.0]$     & $[1.5,\,2.3]$       & \si{\newton\meter\second}\,/\,\si{\radian} \\
\hline
\end{tabular}
\vspace{-0.05cm}
\caption{Domain randomization ranges for intrinsic parameters, $\mathbf{x}_{\text{int}}$.}
\label{table:domainparams}
\end{table}

For deployment, we replace privileged features $\mathbf{z}_{\text{per}}$, $\mathbf{x}_{\text{ext}}$, $\mathbf{z}_{\text{int}}$ with estimates from their corresponding estimators.
The perceptual estimator $e^{\textit{per}}_\phi\!: \mathbf{o}_{\text{cam}}^\text{5} \!\rightarrow \hat{\mathbf{z}}_{\text{per}}$ infers perceptual latent features from a sequence of five belly-mounted camera images. 
To improve robustness to lighting changes and specular reflections, the RGB images are filtered by the deck color in the HSV color space to obtain semantic masks, and randomly perturbed with salt-and-pepper noise or set entirely to zero, as shown in Fig.~\ref{fig:vars}.
The extrinsic estimator $e^{\textit{ext}}_\phi\!: (\mathbf{o}^\text{10}, \hat{\mathbf{z}}_{\text{per}}) \!\rightarrow \hat{\mathbf{x}}_{\text{ext}}$ explicitly infers foot states and deck pose relative to the body, while the intrinsic estimator $e^{\textit{int}}_\phi\!: \mathbf{o}^\text{10} \!\rightarrow \hat{\mathbf{z}}_{\text{int}}$ extracts latent domain features from a 10-observation history.
These estimators are trained by minimizing the distillation loss:
\begin{align}
\mathcal{L}_{\text{DAgger}}(\phi) =& \;
 \mathbb{E}_{\mathcal{D}} \big[
 \left\| \mathbf{z}_{\text{per}} - \hat{\mathbf{z}}_{\text{per}}(\phi) \right\|_2^2 \notag \\
&+ \left\| \mathbf{x}_{\text{ext}} - \hat{\mathbf{x}}_{\text{ext}}(\phi) \right\|_2^2 
 + \left\| \mathbf{z}_{\text{int}} - \hat{\mathbf{z}}_{\text{int}}(\phi) \right\|_2^2
\big],
\label{eq:dagger}
\end{align}
where $\phi$ is the estimators' parameters to be optimized. 
The dataset $\mathcal{D}$ is constructed by iteratively rolling out the policy and annotating the estimators’ observations with supervision.

As listed in TABLE~\ref{tab:mode_rewards}, we define the reward function as  
$\mathcal{R}_{\hspace{-0.1em}M} = \mathcal{R}_{\text{dep}} + \mathcal{R}_{\text{ind}}$,  
where $\mathcal{R}_{\text{dep}} = \sum_{i=1}^7 r_i$ contains mode-dependent terms that adapt the policy to each skateboarding mode $M\hspace{-0.08em}(\phi_t)$.
Specifically, $r_1 $--$r_4$ promote accurate command tracking, body and foot placement, and riding posture, while $r_5$--$r_7$ regulate foot slippage, contact patterns, and clearance. 
The mode-independent component $\mathcal{R}_{\text{ind}} = \sum_{i=8}^{11} r_i$ enforces robot-skateboard alignment ($r_8$), joint smoothness ($r_9$), stable movements ($r_{10}$), and safety ($r_{11}$) across all phases to ensure energy efficiency, robustness, and safe operation. 
This structured design enables the phase-conditioned policy to acquire mode-specific skills while maintaining stable behavior throughout the skateboarding cycle.
While the \textsc{Transition} mode has no dedicated reward terms, maximizing expected returns induces mounting and dismounting behaviors in $\phi \in [1.8\pi, 2\pi) \cup [0, 0.2\pi]$ and $\phi \in [0.8\pi, 1.2\pi]$, respectively, for the periodically upcoming \textsc{Carving} and \textsc{Pushing} modes.

\begin{table}[t]
\centering
\renewcommand{\arraystretch}{1.1} 
\begin{tabular}{
    @{\hspace{6pt}}l@{\hspace{0pt}}
    |l@{\hspace{4pt}}
    |l@{\hspace{4pt}}
    |l@{\hspace{2pt}
    }
}
\hline
\textbf{Net.} & \textbf{Inputs (dimension)} & \textbf{Architecture} & \textbf{Outputs} \\ 
\hline 
$f_\theta^{\textit{critic}}$ \rule{0pt}{2.0ex} 
  & [$\mathbf{o}, \mathbf{x}_{\text{scan}}, \mathbf{x}_{\text{full}}$] (1702) 
  & $\mathcal{F}$(1024, 512, 256, 1) & [$V$] (1) \\ [1.0pt]
\hline 
\rule{0pt}{2.0ex}$f_\theta^{\textit{actor}}$ 
  & [$\mathbf{o}, \mathbf{z}_{\text{per}}, \mathbf{z}_{\text{int}}, \mathbf{x}_{\text{ext}}$] (91) 
  & $\mathcal{F}$(512, 256, 128, 12) & [$\mathbf{a}$] (12) \\
$e^{\textit{scan}}_\theta$ 
  & [$\mathbf{x}_{\text{scan}}$] (64 x 86) 
  & 1D CNN-GRU + [16] & [$\mathbf{z}_{\text{per}}$] (16) \\ 
$e^{\textit{enc}}_\theta$ 
  & [$\mathbf{x}_{\text{int}}$] (35) 
  & [128, 64, 16] & [$\mathbf{z}_{\text{int}}$] (16) \\ [1.0pt]
\hline 
\rule{0pt}{2.2ex}$e^{\textit{per}}_{\phi}$ 
  & [$\mathbf{o}_{\text{cam}}^\text{5}$] (128 x 256 x 5) 
  & 2D\,CNN-GRU + [16] & [$\mathbf{\hat{z}}_{\text{per}}$] (16) \\
$e^{\textit{ext}}_{\phi}$ 
  & [$\mathbf{o}^{\text{10}}, \mathbf{\hat{z}}_{\text{per}}$] (45 x 10, 16) 
  & 1D\,CNN-GRU + [14] & [$\mathbf{\hat{x}}_{\text{ext}}$] (14) \\
$e^{\textit{int}}_{\phi}$ 
  & [$\mathbf{o}^{\text{10}}$] (45 x 10) 
  & 1D\,CNN-GRU + [16] & [$\mathbf{\hat{z}}_{\text{int}}$] (16) \\ [1.0pt]
\hline
\end{tabular}
\vspace{-0.05cm}
\caption{
\textbf{Network Architectures.} 
$\mathcal{F}(\cdot)$ denotes a FiLM-modulated MLP with intermediate and final layer dimensions. [\,$\cdot$\,] indicates a standard MLP. CNN-GRU represents a convolutional encoder followed by a gated recurrent unit with a 32-dimensional hidden state, encoding spatiotemporal inputs. 
}
\label{tab:net_arch}
\vspace{-0.65cm}
\end{table}

\begin{table*}[ht!]
\scriptsize
\centering
\setlength{\tabcolsep}{2pt}
\renewcommand{\arraystretch}{0.2}
\caption{
Reward Composition for Skateboard-Riding Skills: $\mathcal{R}_{\hspace{-0.1em}M} = \mathcal{R}_{\text{dep}} + \mathcal{R}_{\text{ind}}$. (For a more detailed description, please refer to Sec.~\ref{sec:arch}.)
}
\vspace{-1mm}
\begin{tabularx}{\textwidth}{
                        >{\arraybackslash}p{0.13\textwidth}|
                        >{\centering\arraybackslash}p{0.13\textwidth}|
                        >{\centering\arraybackslash}p{0.29\textwidth}|
                        >{\centering\arraybackslash}p{0.42\textwidth}
                        }
\toprule
\multicolumn{4}{c}{\textbf{Mode-Dependent Rewards} $\mathcal{R}_{\text{dep}} = \sum_{i=1}^7 r_i$} \\
\midrule
\multirow{2}{*}{$\;\;\;\;\,\,$\textbf{Reward Term}\raisebox{0.5mm}} &
\multirow{2}{*}{\centering\textbf{Formulation}\raisebox{0.5mm}} &
\multicolumn{2}{c}{\textbf{Error Term $\varepsilon_i (i = 0, 1, \cdots, 7)$}} \\
\cmidrule(lr){3-4}
& & $M\hspace{-0.08em}(\phi_t)=\textsc{Carving}$ & $M\hspace{-0.08em}(\phi_t)=\textsc{Pushing}$ \\
\midrule

$\hspace{2mm}r_1:$ Command&
$5.0\exp(-\varepsilon_1/0.3)$ &
$|c_{\omega}\!-\omega^{S}_{D,z}|$ &
$0.6 \, |c_{v} \!-\mathrm{v}^{\mathcal{S}}_{D,x}| + 0.2 \, |\mathrm{v}^{\mathcal{S}}_{D,y}| + 0.2 \, |\omega^{\mathcal{S}}_{D,z}|$ \\
\midrule
$\hspace{2mm}r_2:$ Body Position&
$2.0\exp(-\varepsilon_2/0.2)$&
$ |\mathbf{p}^{\mathcal{S}}_{B, xz}| + 0.2 \, |\mathrm{p}^{\mathcal{S}}_{B, y}| + |\mathrm{p}^{\mathcal{S}}_{B, z} \! - \mathrm{p}^{\text{carving}}_{B, z}|$  &
$\| \mathbf{p}^{\mathcal{S}}_{B} - \mathbf{p}^{\text{pushing}}_{B} \|_{2}$ \\
\midrule
\multirow{2}{*}{$\hspace{2mm}r_3:$ Foot Position\raisebox{0.5mm}} &
\multirow{2}{*}{$2.0\exp(-\varepsilon_3/0.3)$\raisebox{0.5mm}} &
\multirow{2}{*}{$\sum_{i \in \mathcal{F}_{\text{all}}}|\mathbf{p}^{\mathcal{S}}_{F_i,xy} \!- \mathbf{p}^\text{carving}_{F_i, xy}|$} &
$\sum_{i \in \mathcal{F}_{\text{right}}} \!|\mathbf{p}^{\mathcal{S}}_{F_i,xy} \!- \mathbf{p}^\text{pushing}_{F_i, xy}| \;+$ \\[1.5pt]
& & & $\sum_{j \in \mathcal{F}_{\text{left}}} \! 0.2 \, |\mathrm{p}^{\mathcal{S}}_{F_j,x} - \mathrm{p}^\text{pushing}_{F_j, x}| + 0.8 \, |\mathrm{p}^{\mathcal{S}}_{F_j,y} \!- \mathrm{p}^\text{pushing}_{F_j, y}|$ \\
\midrule
$\hspace{2mm}r_4:$ Riding Posture&
$2.0\exp(-\varepsilon_4/0.4)$ &
$\|\boldsymbol{q} - \boldsymbol{q}^{\text{carving}}\|_1$ &
$0.7 \sum_{i \in \mathcal{J}_{\text{right}}} \! |q_{i}-q^{\text{pushing}}_{i}| + 0.3 \sum_{j \in \mathcal{J}_{\text{left}}} \! |q_{j}-q^{\text{pushing}}_{j}|$ \\
\midrule
$\hspace{2mm}r_5:$ Foot Slip&
$1.0\exp(-\varepsilon_5/0.4)$ &
$\sum_{i \in \mathcal{F}_{\text{all}}} \!\| \mathbf{v}^{\mathcal{D}}_{F_i, xy}\|_2 \mathbbm{1}_{\{ c_{i}=1 \}}$& 
$\sum_{i \in \mathcal{F}_{\text{right}}} \!\| \mathbf{v}^{\mathcal{D}}_{F_i, xy}\|_2 \mathbbm{1}_{\{\! c_{i}=1 \!\}} + \sum_{i \in \mathcal{F}_{\text{left}}} \!\| \mathbf{v}^{\mathcal{W}}_{F_i, xy}\|_2 \mathbbm{1}_{\{\! c_{i}=1 \!\}}$ \\
\midrule
$\hspace{2mm}r_6:$ Contact Pattern&
$- 2.0\,\varepsilon_6$ &
$4 - \sum_{i \in \mathcal{F}_{\text{all}}} \!\mathbbm{1}_{\{\! c_{i}=1 \!\}}$& 
$2 - \sum_{i \in \mathcal{F}_{\text{right}}} \!\mathbbm{1}_{\{\! c_{i}=1 \!\}}$ \\
\midrule
$\hspace{2mm}r_7:$ Foot Clearance&
$- 0.4\,\varepsilon_7$ &
$0$& 
$\sum_{i \in \mathcal{F}_{\text{left}}} \! \max(0.1 - p^{\mathcal{W}}_{F_i, z}, 0.0) \mathbbm{1}_{\{\! c_{i}=0 \!\}} $ \\

\midrule
\multicolumn{4}{c}{\textbf{Mode-Independent Rewards} $\mathcal{R}_{\text{ind}} = \sum_{i=8}^{11} r_i$} \\ 
\midrule

$\hspace{2mm}r_8:$ Alignment&
\multicolumn{3}{c}{$
2.0 \exp(-\| \boldsymbol{\theta}^{\mathcal{S}}_{B, xy} \|_2 / 0.5)
+ 2.0\exp(-|\theta^{\mathcal{S}}_{B, z}|/0.2)
+ \exp(-\|\mathbf{v}^{\mathcal{S}}_{B, xy}\|_2/0.3)
+ 2.0\exp(-\sum_{i \in \mathcal{F}_{\text{all}}} \! \|\boldsymbol{\theta}^{\mathcal{S}}_{F_i, xz}\| / 0.9)
$} \\
\midrule
$\hspace{2mm}r_9:$ Smoothness&
\multicolumn{3}{c}{
$
-5e\text{-}6\| \mathbf{a} - \mathbf{a}_{t-1} \|_2 
-1e\text{-}6 \| \boldsymbol{\Dot{q}} \|_2 
-1e\text{-}7 \| \boldsymbol{\Ddot{q}} \|_2 
-4e\text{-}7 \| \boldsymbol{\tau_q} \|_2 
-1e\text{-}6 \sum_{j \in \mathcal{J}_{\text{all}}} \!\max( \tau_{q}[j] \Dot{q}[j], 0.0)
$}\\
\midrule

$\;r_{10}:$ Stabilization&
\multicolumn{3}{c}{
$
-1e\text{-}6 \| \dot{\mathbf{v}}^{\mathcal{W}}_{B} \|_2
-7e\text{-}2\| \boldsymbol{\omega}^{\mathcal{B}}_{B, xy} \|_2 
-2.0 | \mathrm{v}^{\mathcal{W}}_{B, z} | 
-1e\text{-}5 \| \dot{\mathbf{v}}^{\mathcal{W}}_{D} \|_2
-1e\text{-}4 |\omega^S_{D, y}|
-1e\text{-}4 \|\dot{\boldsymbol{\omega}}^S_{D}\|_2
$}\\
\midrule

$\;r_{11}:$ Safety&
\multicolumn{3}{c}{
$
- 2.0 \sum_{i \in \mathcal{F}_{\text{all}}} \!\mathbbm{1}_{\{F_i \text{is on edge}\}} 
- 1.0 \sum_{i \in \mathcal{F}_{\text{all}}} \max(\|\mathbf{f}_{F_i}\|_2 - 80.0, 0.0)
- 2.5 \,\mathbbm{1}_{\{\text{collision}\}}
- 6.0 \,\mathbbm{1}_{\{\text{termination}\}}
- 2.0 \,\mathbbm{1}_{\{\text{joint limit}\}}
$}\\

\bottomrule
\end{tabularx}
\label{tab:mode_rewards}
\vspace{-0.5cm}
\end{table*}

\subsection{Implementation Details} \label{sec:details}

We employed Isaac Gym~\cite{makoviychuk2021isaac} to collect data using 4,096 parallel environments, each with a Unitree Go1 robot~\cite{Unitree_Go1} and a skateboard measuring $0.69 \times 0.27 \times 0.13$\,\si{\meter} with a $0.43$\,\si{\meter} wheelbase.
We set the steering parameters in Eq.~(\ref{eq:steering}) to $\gamma_F^1 \!=\! 1.0$, $\gamma_F^2 \!= \!1.12$, $\gamma_R^1 \!=\! 0.7$, and $\gamma_R^2 \!=\! 0.9$, yielding realistic behavior where a $10^\circ$ deck roll induces about $11.0^\circ$ front and $6.2^\circ$ rear axle yaw.
We tuned the front truck to favor responsiveness for a balance between stability and agility.

To enhance domain adaptability, we randomized the intrinsic parameters $\mathbf{x}_{\text{int}}$ of both the robot and skateboard within the ranges listed in TABLE~\ref{table:domainparams}. The overall architecture and network configuration are illustrated in Fig.\ref{fig:overall} and TABLE~\ref{tab:net_arch}. To improve robustness to external perturbations and sudden command changes, we applied random impulses to the body and deck every \SI{3}{\second} and resampled the command $\boldsymbol{c}$ every \SI{5}{\second}.

We used a stochastic policy $\pi_\theta$ that samples actions from a diagonal Gaussian, with the mean predicted by the actor network $f^{\text{actor}}_\theta$ and learnable standard deviations $\boldsymbol{\theta_{\text{std}}} \in \mathbb{R}^{12}$.
The policy converged in about \SI{10}{\hour} on an RTX~4090 GPU with an Intel i9-9900K CPU, followed by \SI{5}{\hour} for distillation.

    \section{Experimental Results}
\label{sec:4}
We demonstrate the effectiveness of the proposed Phase-Aware Policy Learning (\textit{PAPL}) framework through a series of experiments. 
First, we verify the intrinsic multi-modality of skateboarding motions, motivating the adoption of a phase-aware modulation. 
Next, we evaluate the command-tracking accuracy and conduct ablation studies to quantify the contribution of individual components. 
Finally, we compare energy efficiency with other locomotion baselines and demonstrate sim-to-real transferability on a physical skateboard platform.

\subsection{Multi-Modality of Skateboarding Motions} \label{result:verification}
Skateboarding inherently involves multi-modal behaviors across distinct phases. 
To verify this property, we visualized the distributions of action, observation, and critic-value data collected over \SI{80}{\second} of execution using a proficient skateboard-riding policy. 
To collect a range of motion trajectories, we resampled the command $\boldsymbol{c}$ every \SI{10}{\second} within $c_v \in [0.0, 1.5]$ and $c_w \in [-0.5, 0.5]$, with a fixed phase period of $T_\phi =\! \SI{10}{\second}$.

\begin{figure}[t!]
    \centering 
    \includegraphics[width=\columnwidth]{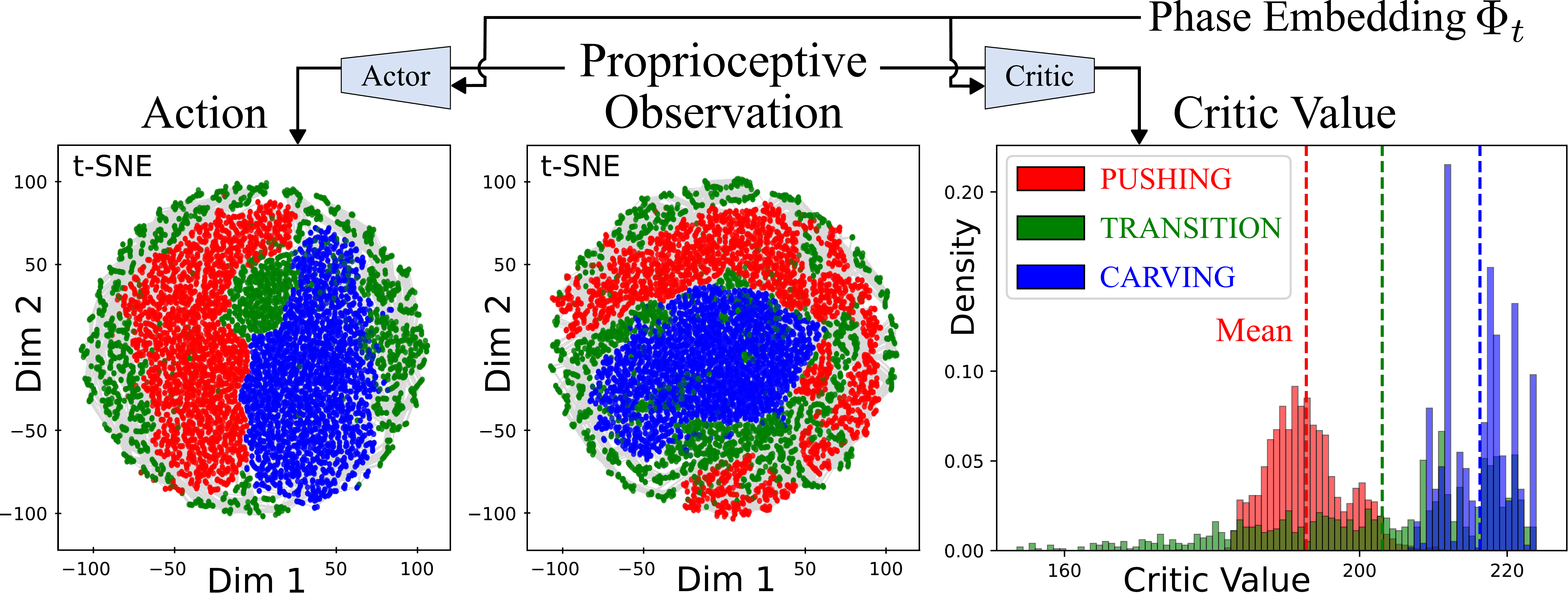}
    \caption{
Visualization of action $\boldsymbol{a}$, proprioceptive observation $\mathbf{o}_{\text{prop}}$, and critic-value $V$ distributions over \SI{80}{\second} of skateboard riding. 
Action and observation data are projected using t-SNE~\cite{maaten2008visualizing}, and critic values are visualized as histograms. Data points are color-coded by the corresponding mode $M\hspace{-0.08em}(\phi_t)$ at the time of collection. Further details are provided in Sec.~\ref{result:verification}.
    }
    \label{fig:tsne}
    \vspace{-0.6cm}
\end{figure}

As illustrated in Fig.~\ref{fig:tsne}, t-SNE projections of action and observation data show mode-dependent clustering: \textsc{Pushing} (red) and \textsc{Carving} (blue) occupy separated regions, while \textsc{Transition} (green) lies in between. While dimensionality reduction could exaggerate separation, the observed structure indicates that the actor needs to address phase-specific variations in its input–output distributions. On the critic side, the value histogram also exhibits distinct distributions reflecting mode-dependent rewards $\mathcal{R}_{\hspace{-0.1em}M}$, listed in TABLE~\ref{tab:mode_rewards}. These findings motivate the use of phase-conditioned modulation to effectively address different modalities using a unified policy.

\subsection{Evaluation of Skateboarding Performance}
We evaluated skateboarding performance and the contribution of each \textit{PAPL} component by measuring time-averaged command-tracking errors of our policy and its ablated variants.
We defined the error term as $|c_{v} - v^{\mathcal{S}}_{D,x}|\mathbbm{1}_{\{M=\textsc{Pushing}\}} + |c_{\omega} - \omega^{S}_{D,z}|\mathbbm{1}_{\{M=\textsc{Carving}\}}$ and computed it over the command set sampled from $c_v \in [0.0, 1.5]$ and $c_\omega \in [-0.5, 0.5]$ at 0.05 resolution.
For each command, ten environments were initialized with intrinsic properties drawn from the test ranges in TABLE~\ref{table:domainparams}. Robots executed $c_v$ for \SI{5}{\second} and $c_\omega$ for another \SI{5}{\second} ($T_\phi=\SI{10}{\second}$), repeated over a \SI{40}{\second} horizon, while random body perturbations are applied every \SI{4}{\second} to assess robustness.

\begin{figure*}[t!]
    \centering 
    \includegraphics[width=2\columnwidth]{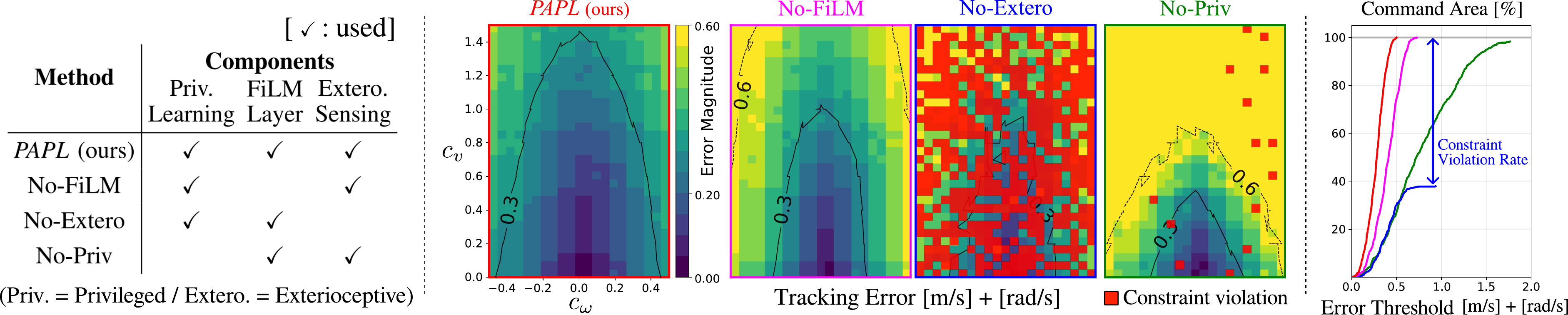}
    \caption{
\textbf{Left}: Component configurations of the proposed \textit{PAPL} framework and ablated variants.
\textbf{Middle}: Tracking error heatmaps with contours, where darker regions indicate lower errors. Contours represent iso-error boundaries, while red regions denote constraint violations when the robot either overturned or deviated more than \SI{0.5}{\meter} from the board.
\textbf{Right}: Command-area curves showing the percentage of commands tracked within given error thresholds. 
    }
    \label{fig:cmd}
    \vspace{-0.4cm}
\end{figure*}

Fig.~\ref{fig:cmd} shows tracking-error heatmaps and corresponding command-area curves~\cite{margolis2024rapid}, where the command area denotes the portion of the command space within a specified error threshold. 
Trials in which the robot overturned or deviated more than \SI{0.5}{\meter} from the board were treated as constraint violations and excluded from the command-area calculation.
\textit{PAPL} presents the largest region with tracking error below 0.3 and no constraint violations, demonstrating robustness across broad command spaces and under external disturbances.
By contrast, the No-FiLM variant, which employs standard MLP backbones for both actor and critic, exhibits a narrower region with higher errors. This empirically demonstrates that the inductive bias introduced by the FiLM architecture is crucial to address the multi-modality inherent in skateboarding motions. 
Meanwhile, the No-Extero variant incurs frequent constraint violations, underscoring the critical role of exteroceptive sensing. In particular, the policy often fails to correct external perturbations or persistent drifts, which eventually leads to collapse.
The No-Priv variant fails to acquire effective motions and, in most cases, converges to a standing-still behavior to avoid penalties.
We also evaluated the Mixture-of-Experts (MoE) MLP as the actor backbone in place of the FiLM MLP, but it failed to achieve effective skateboarding behaviors as the pushing-phase expert collapsed without producing meaningful actions. This failure was likely due to phase-specific isolation, limited information sharing among experts, and the large parameter count, which under limited GPU memory restricted training to fewer environments, reducing data collection and exploration efficiency and leading to trivial behaviors such as remaining inactive.

\begin{figure}[t!]
    \centering 
    \includegraphics[width=\columnwidth]{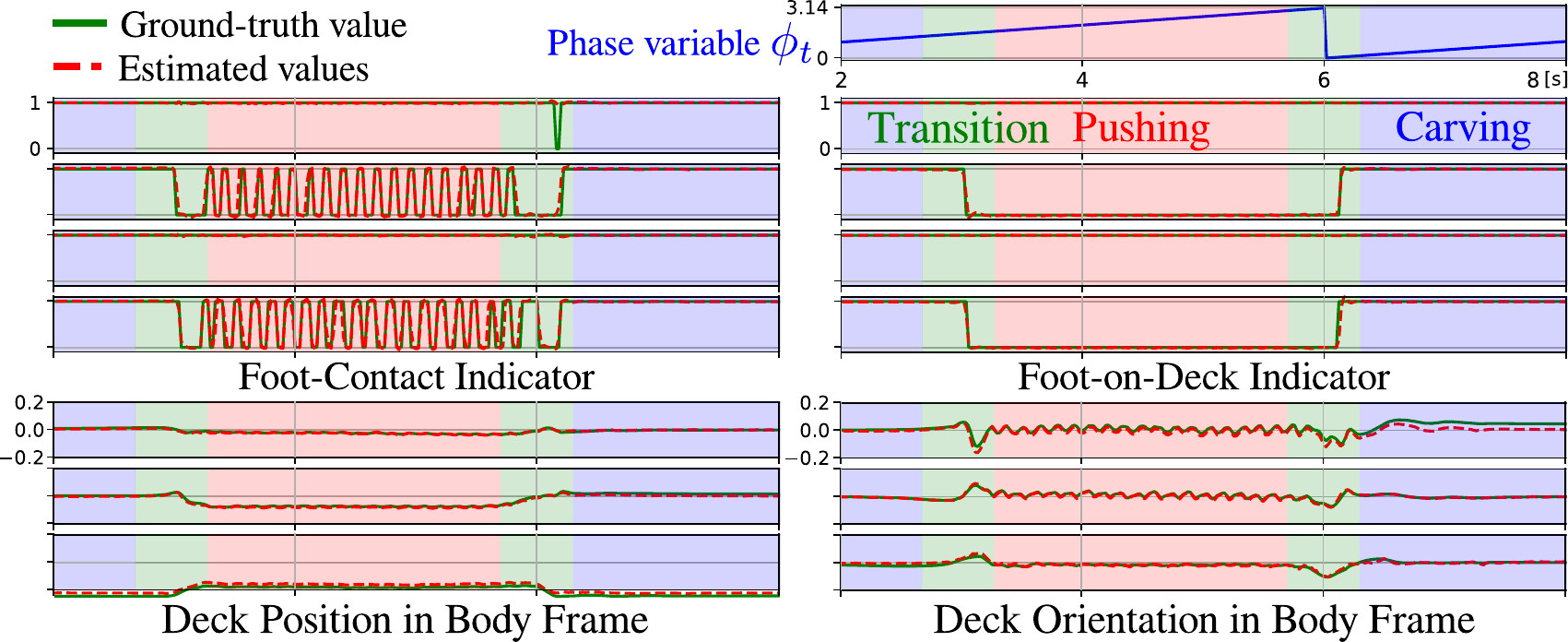}
    \caption{
Estimation performance of the extrinsic estimator across skateboarding phases. It accurately infers foot-contact and foot-on-deck indicators, as well as deck position and orientation in the body frame, closely matching ground truth values during skateboarding at $c_v=\SI{0.8}{\meter/\second}$ with $T_\phi=\SI{6}{\second}$.
    }
    \label{fig:exp_est}
    \vspace{-0.6cm}
\end{figure}

The command-area curves show that \textit{PAPL} attains broader command-space coverage under strict error thresholds, highlighting the complementary roles of its components.
Fig.~\ref{fig:exp_est} further shows the extrinsic estimator's accuracy in inferring privileged extrinsic states $\mathbf{x}_{\text{ext}}$ from observation histories; additional results are available in the supplementary video.

\subsection{Power Consumption Analysis} \label{sec:power}
\begin{figure}[t]
    \centering 
    \includegraphics[width=\columnwidth]{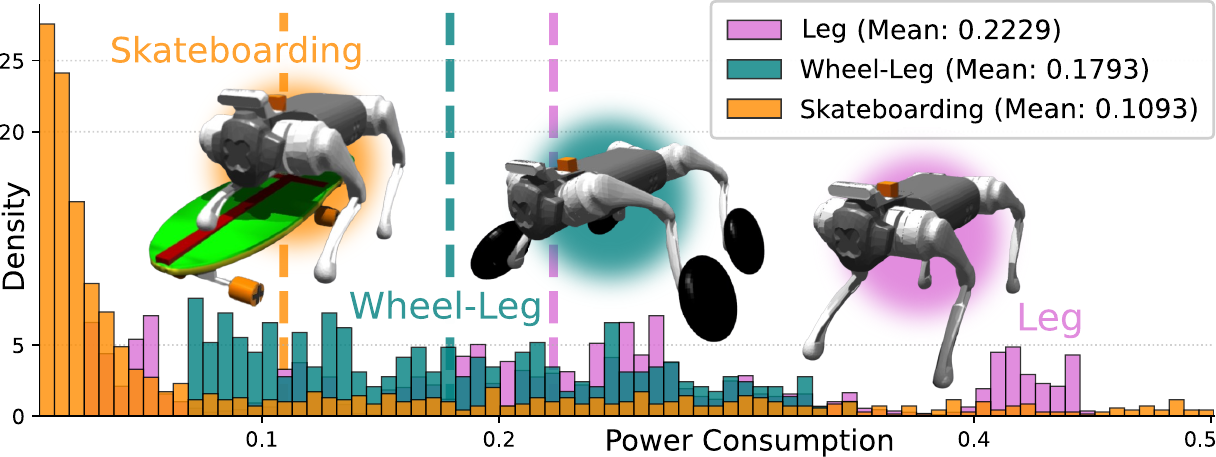}
    \caption{
Consumed motor power distributions for three locomotion strategies over a \SI{30}{\meter} traversal. 
Vertical dashed lines denote the mean values. 
    }
    \label{fig:energy}
    \vspace{-0.7cm}
\end{figure}

To compare locomotion efficiency, we measured normalized motor power~\cite{yoon2025enhancing} required to reach a target \SI{30}{\meter} ahead on flat ground.
For the wheel-legged and legged baselines, the forward velocity command was fixed at \SI{1.5}{\meter/\second}, and heading was stabilized using yaw-rate commands from a PD controller on heading error.
For the skateboarding, we strategically modulated the phase clock $\phi_t$: propulsion was activated only when the forward velocity dropped below \SI{0.7}{\meter/\second}, otherwise the clock was halted in the middle of carving phases to focus solely on steering. 
Although this comparison was conducted on flat terrain with moderate friction favorable to skateboarding, Fig.~\ref{fig:energy} shows that skateboarding consumes less power than the other two locomotion strategies. The histogram of skateboarding exhibits a long-tail distribution, reflecting its unique characteristics: steering maneuvers require relatively little energy, whereas motions such as pushing off and mounting need much higher power.

\subsection{Real-World Experiments}
\begin{figure}[t!]
    \centering 
    \includegraphics[width=\columnwidth]{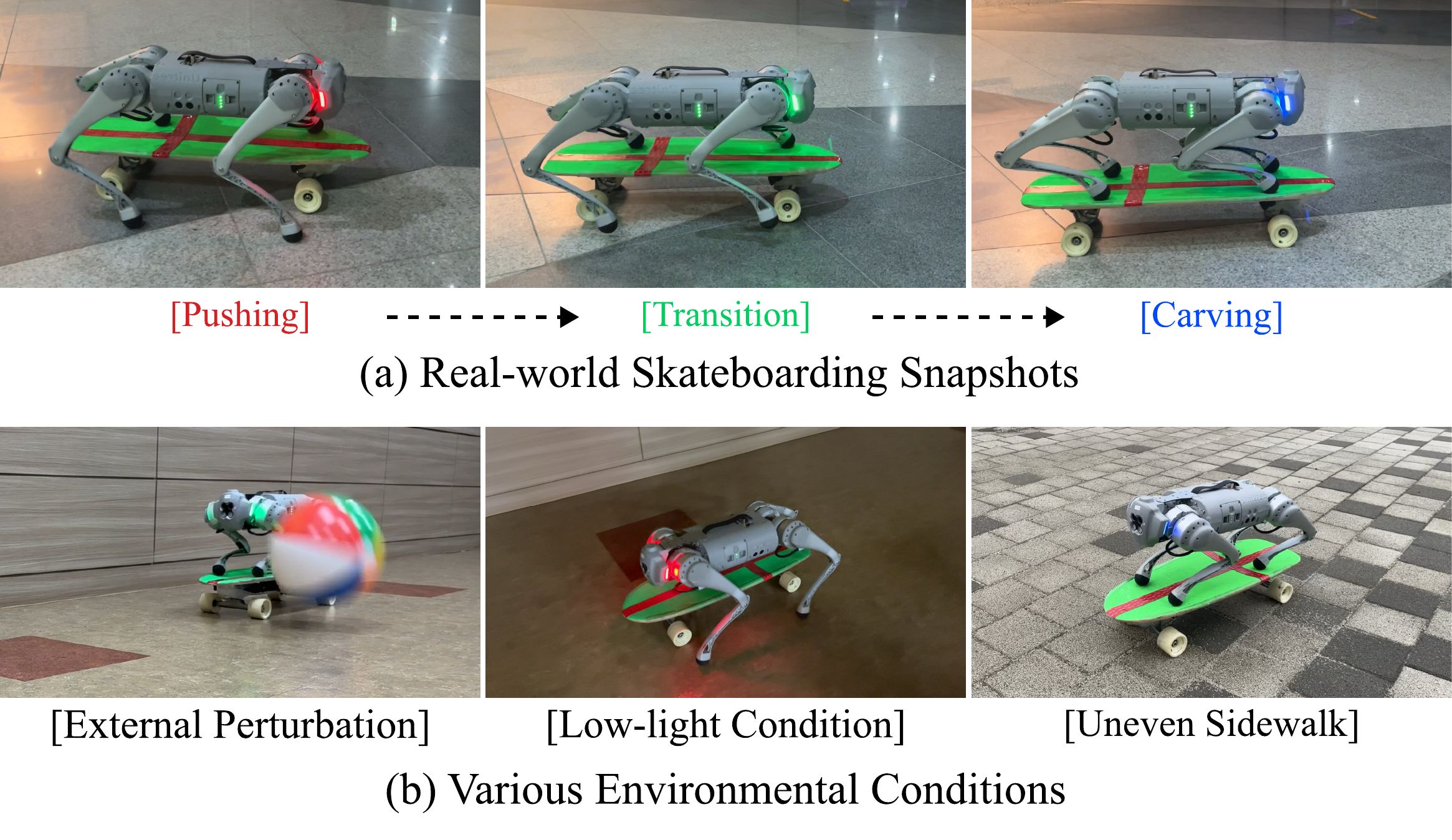}
    \vspace{-0.6cm}
    \caption{
    Real-world demonstrations of quadruped skateboarding. (a) Snapshots captured at different modes. (b) Various experimental conditions: external perturbation, low-light settings, and uneven sidewalks.
    }
    \label{fig:real}
    \vspace{-0.6cm}
\end{figure}

We applied the proposed Phase-Aware Policy Learning (\textit{PAPL}) framework to a Unitree Go1 robot riding a physical skateboard to examine its real-world applicability. 
As shown in Fig.~\ref{fig:real}-(a), the robot reproduced pushing, transition, and carving behaviors via zero-shot transfer from simulation. 
We further conducted trials under diverse conditions, including external perturbations, low-light environments, and uneven sidewalks (Fig.~\ref{fig:real}-(b)), confirming that the policy can be deployed without collapse. 
These experiments validate the feasibility of our approach in hardware; for an intuitive understanding, please refer to the supplementary video.

    \section{Conclusion}
\label{sec:5}
In this work, we introduced the Phase-Aware Policy Learning (\textit{PAPL}) framework to endow quadruped robots with skateboarding capability for efficient transportation. 
By integrating a phase-conditioned RL formulation with FiLM-based layer modulation, asymmetric privileged learning, and exteroceptive sensing, \textit{PAPL} achieves robust and agile skateboarding behaviors, including steering via deck tilting.
Simulation results demonstrated moderate command tracking performance of robots using skateboards, improved energy efficiency over other locomotion baselines on the flat ground, and the complementary role of each component through ablation studies. 
Real-world experiments further demonstrated zero-shot transfer from simulation, where the robot successfully operated under diverse conditions including external perturbations, low-light settings, and uneven terrains.

As future work, we plan to integrate a high-level navigation policy that generates velocity commands and modulates the phase clock in response to surrounding environmental conditions—particularly for timely mounting and propulsion—and to incorporate front-facing perception modules for velocity estimation and navigation command generation.

    {
        \small
        \bibliographystyle{ieee_format/ieee}
        \bibliography{./ref}
    }

\end{document}